\documentclass[sigconf]{acmart}
\usepackage[utf8]{inputenc}

\renewcommand\footnotetextcopyrightpermission[1]{} 
\usepackage{booktabs} 

\usepackage{color}
\usepackage{todonotes} 

\usepackage{rotating,dblfloatfix,graphicx,caption,subcaption,amsmath,amssymb,tabularx,array,multirow,wrapfig,setspace,etoolbox,eqnarray,multicol,csquotes,enumitem}
\usepackage{epstopdf}
\usepackage{listings,booktabs}
\usepackage{microtype}
\usepackage{siunitx}
\usepackage[framemethod=TikZ]{mdframed}

\newcommand\customArrow{\scalebox{2.0}{$\boldsymbol{\downarrow}$}}

\setcopyright{none}

\begin{document}
\title{Which Knowledge Graph Is Best for Me?} 
\subtitle{``Linked Data Quality of DBpedia, Freebase, OpenCyc, Wikidata, and YAGO'' in a Nutshell}

\author{Michael Färber}
\orcid{0000-0001-5458-8645}
\affiliation{%
  \institution{University of Freiburg}
  \streetaddress{}
  \city{Freiburg} 
  \state{Germany} 
  \postcode{}
}
\email{michael.faerber@cs.uni-freiburg.de}

\author{Achim Rettinger}
\affiliation{%
  \institution{Karlsruhe Institute of Technology (KIT)}
  \city{Karlsruhe} 
  \country{Germany}}
\email{rettinger@kit.edu}

\renewcommand{\shortauthors}{M. Färber and A. Rettinger}

\begin{abstract}
In recent years, DBpedia, Freebase, OpenCyc, Wikidata, and YAGO have been published as noteworthy large, cross-domain, and freely available knowledge graphs. 
Although extensively in use, these knowledge graphs are hard to compare against each other in a given setting. 
Thus, it is a challenge for researchers and developers to pick the best knowledge graph for their individual needs.
In our recent survey \cite{FaerberSWJ2016}, we devised and applied data quality criteria to the above-mentioned knowledge graphs. Furthermore, we proposed a framework for finding the most suitable knowledge graph for a given setting. 
With this paper we intend to ease the access to our in-depth survey by presenting simplified rules that map individual data quality requirements to specific knowledge graphs.
However, this paper does not intend to replace the decision-support framework introduced in \cite{FaerberSWJ2016}. For an informed decision on which KG is best for you we still refer to our in-depth survey.
\end{abstract}


\maketitle

\section{Introduction}

Did you ever have to make a quick decision on which publicly available knowledge graph (KG) to use for a given task? This paper provides you with a list of simplified rules of thumb which recommend a KG given individual data quality requirements.
In order to generate such rules a systematic overview of KGs is needed, similar to the ``Michelin guide to knowledge representation'' \cite{Markman2013}. We laid this groundwork in our previous article \cite{FaerberSWJ2016} where we provide an in-depth analysis of KGs and propose an extensive KG recommendation framework. 
There, we limited ourselves to the  KGs DBpedia, Freebase, OpenCyc, Wikidata, and YAGO,\footnote{We considered the KGs in their versions available in April 2016.} as they are freely accessible and freely usable from within the Linked Open Data (LOD) cloud, and as they cover general knowledge, not selected domains only. Both aspects make these KGs widely applicable.\footnote{An indicator for that statement is the high, and still steadily increasing number of publications referring to the considered KGs: According to Google Scholar, about 26k/21k/4k/5k/46k publications mention ``DBpedia''/``Freebase''/``OpenCyc''/``Wikidata''/``YAGO'' on Sep 25, 2018.}

This paper intends to provide a simplified summary of our in-depth analysis in \cite{FaerberSWJ2016}. This includes  
(i) an overview how data quality can be measured when it comes to KGs (see Section \ref{sec:data-quality-criteria} concerning data quality criteria for KGs) and
(ii) a crisp overview of the KGs DBpedia, Freebase, OpenCyc, Wikidata, and YAGO using key statistics (see Section \ref{sec:key-statistics}). 

Moreover, in our survey \cite{FaerberSWJ2016}, we applied the developed data quality criteria to these KGs. Based on those previous findings and at the risk of oversimplification, we created rules of thumbs in the form ``Pick KG X if requirement Y holds'' (see Section~\ref{sec:applying-metrics}) for this paper. While these help to get a rough idea which KG might be best for you, we still recommend to use our full KG recommendation framework for making  thorough decisions. This framework is outlined in Section~\ref{sec:framework} and presented in detail in~\cite{FaerberSWJ2016}.

\begin{table*}[tb]
 \centering
 \begin{small}
  \caption{Summary of key statistics.}
   \label{tab:summary-key-figures}%
\begin{tabular}{lS[table-parse-only,table-number-alignment=right]S[table-parse-only,table-number-alignment=right]S[table-parse-only,table-number-alignment=right]S[table-parse-only,table-number-alignment=right]S[table-parse-only,table-number-alignment=right]}
\toprule
                                            & {DBpedia}   & {Freebase}   & {OpenCyc} & {Wikidata}  & {YAGO}       \\ 
                                            \midrule
Number of triples   & 411885960 & 3124791156 & 2412520 & 748530833 & 1001461792 \\
Number of classes                               & 736       & 53092      & 116822  & 302280    & 569751     \\
Number of relations                         & 2 819     & 70902      & 18028   & 1874      & 106        \\
No. of unique predicates                            & 60231     & 784977     & 165     & 4839      & 88736      \\
Number of entities                          & 4298433   & 49947799   & 41029   & 18697897  & 5130031    \\
Number of instances                          & 20764283  & 115880761  & 242383  & 142213806 & 12291250   \\
Avg. number of entities per class                           & 5840.3    & 940.8      & 0.35    & 61.9      & 9.0          \\
No. of unique subjects 
       & 31 391 413  & 125 144 313  & 261 097  & 142 278 154 & 331806927  \\
No. of unique non-literals in object position  & 83284634  & 189466866  & 423432  & 101745685 & 17438196   \\
No. of unique literals in object position                         & 161398382 & 1782723759 & 1081818 & 308144682 & 682313508 \\
\bottomrule
\end{tabular}
\end{small}
\end{table*}

\section{Data Quality Criteria for Knowledge Graphs}
\label{sec:data-quality-criteria}

Based on existing works on data quality in general (see, in particular, the data quality evaluation framework of Wang et al.~\cite{Wang1995}) and on data quality of Linked Data in particular (see \cite{Bizer2007} and \cite{Zaveri2014}), we define 
11 \textit{data quality dimensions} for assessing KGs:
\begin{itemize}
    \item Accuracy
    \item Trustworthiness
    \item Consistency
    \item Relevancy
    \item Completeness
    \item Timeliness
    \item Ease of understanding
    \item Interoperability
    \item Accessibility
    \item License
    \item Interlinking
\end{itemize}
Each of the dimensions is a perspective how data quality can be viewed, and each dimension is associated with one or several \textit{data quality criteria} (e.g., ``semantic validity of triples''), which specify different aspects of the data quality dimension. 
In order to measure the degree to which a certain data quality criterion (and, hence, data quality dimension) is fulfilled for a given KG, each criterion is formalized and expressed in terms of a function, which we call the \textit{data quality metric}. 
In case of the criterion ``semantic validity of triples'', 
this metric could be the degree to which all considered statements are semantically correct (assuming that all entities and relations are both in the KG and in a ground truth). 
The values of all data quality metrics, weighted by the user, can then be used for judging the KGs for a concrete setting (see Section~\ref{sec:applying-metrics} and Section~\ref{sec:framework}).

\section{Key Statistics of Knowledge Graphs}
\label{sec:key-statistics}

We statistically compare the RDF KGs DBpedia, Freebase, OpenCyc, Wikidata, and YAGO 
(cf. Table~\ref{tab:summary-key-figures}) and present here our essential findings: 
\begin{enumerate}
\item \textit{Triples}:
All considered KGs are very large. Freebase is the largest KG in terms of number of triples, while OpenCyc is the smallest KG. 
We notice a correlation between the way of building up a KG and the size of the KG: automatically created KGs are typically larger, as the burdens of integrating new knowledge become lower.
Datasets which have been imported into the KGs, such as MusicBrainz into Freebase, have a huge impact on the number of triples and on the number of facts in the KG. 
Also the way of modeling data has a great impact on the number of triples. For instance, if n-ary relations are expressed in N-Triples format (as in case of Wikidata), many intermediate nodes need to be modeled, leading to many additional triples compared to plain statements. Last but not least, the number of supported languages influences the number of triples. 

\item \textit{Classes}: 
The number of classes is highly varying among the KGs, ranging from 736 (DBpedia) up to 300K (Wikidata) and 570K (YAGO). Despite its high number of classes, YAGO contains in relative terms the most classes which are actually used (i.e., classes with at least one instance). This can be traced back to the fact that heuristics are used for selecting appropriate Wikipedia categories as classes for YAGO. Wikidata, in contrast, contains many classes, but out of them only a small fraction is actually used on instance level. Note, however, that this is not necessarily a burden.  
\item \textit{Domains}: 
Although all considered KGs are specified as crossdomain, the domains are not equally distributed in the KGs. Also the domain coverage among the KGs differs considerably. Which domains are well represented heavily depends on which datasets have been integrated into the KGs. MusicBrainz facts had been imported into Freebase, leading to a strong knowledge representation (77\%) in the domain of \textit{media} in Freebase. In DBpedia and YAGO, the domain \textit{people} is the largest, likely due to Wikipedia as data source. 
\item \textit{Relations and Predicates}: Many relations are rarely used in the KGs: Only 5\% of the Freebase relations are used more than 500 times and about 70\% are not used at all. In DBpedia, half of the relations of the DBpedia ontology are not used at all and only a quarter of the relations is used more than 500 times. For OpenCyc, 99.2\% of the relations are not used. We assume that they are used only within Cyc, the commercial version of OpenCyc.
\item \textit{Instances and Entities}: Freebase contains by far the highest number of entities. Wikidata exposes relatively many instances in comparison to the entities (in the sense of instances which represent real world objects), as each statement is instantiated leading to around 74M instances which are not entities. 
\item \textit{Subjects and Objects}: 
YAGO provides the highest number of unique subjects among the KGs and also the highest ratio of the number of unique subjects to the number of unique objects. 
This is due to the fact that N-Quad representations need to be expressed via intermedium nodes
and that YAGO 
is concentrated on classes which are linked by entities and other classes,
but which do not provide outlinks. 
DBpedia exhibits more unique objects than unique subjects, since it contains many \texttt{owl:sameAs} statements to external entities. 
\end{enumerate}

\section{Applying Data Quality Metrics to Knowledge Graphs}
\label{sec:applying-metrics}

\begin{figure}[tb]
 \centering
 \includegraphics[width=0.92\linewidth]{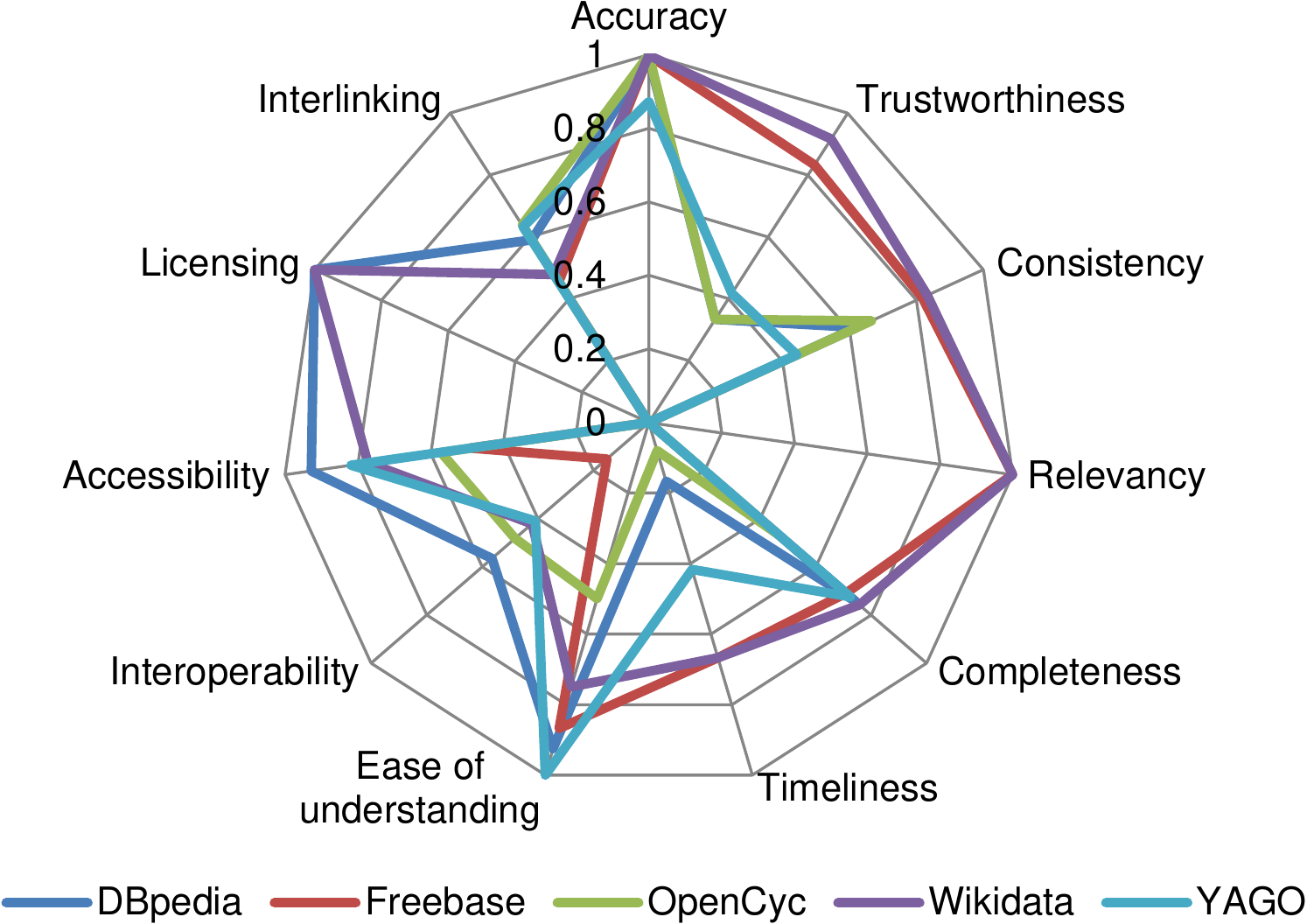} 
 \caption{Results of our data quality assessment, gained by averaging the corresponding data quality metric scores for each data quality dimension.}
  \label{fig:radar-chart}
\end{figure}

When applying our proposed data quality metrics 
to the considered KGs, first of all we obtain scores for each KG with regard to the different data quality metrics. These metrics, each corresponding to a data quality criterion, can be grouped by data quality dimensions. Figure~\ref{fig:radar-chart} shows 
the scores of all data quality dimensions for each KG, each calculated as average over the corresponding data quality metric values. 
We also explore in more detail the reasons for the obtained values and refer in this regard to our article~\cite{FaerberSWJ2016}.

In the following, we use the identified KG characteristics 
to give some general advice
when to use which KG. 
Note that this list of items only highlights some selected features of the respective KGs. Note also that this list is meant to serve as a rough orientation instead of a thorough recommendation. For a more nuanced discussion and selection advice see Section~\ref{sec:framework} and our main article~\cite{FaerberSWJ2016}.

\paragraph{Pick DBpedia...}
\begin{itemize}
 \item if Wikipedia's infoboxes should be exploited explicitly; 
 \item if relations should be covered well, not so much classes; 
 \item if predicates should on average be very frequently used by all instances; 
 \item if descriptive URIs are desired; 
 \item if n-ary relations are to some degree acceptable;
 \item if external vocabulary should be used to a high degree; 
 \item if rather classes than relations should have equivalent-statements to entries in other data sources;
 \item if many instances should have \texttt{owl:sameAs} links to entries in other data sources besides Wikipedia; 
 \item if it is not that important whether linked RDF documents are not accessible any more;
\end{itemize}

\paragraph{Pick Freebase...}
\begin{itemize}
 \item if the possibility to store unknown and empty values should be given; 
 \item if classes may belong to various domains; 
 \item if predicates should on average be very frequently used by all instances; 
 \item if the period in which statements are valid need to be represented; 
 \item if the modification date of statements need to be kept; 
 \item if labels should be available in hundreds of languages; 
 \item if n-ary relations are acceptable; 
 \item if a SPARQL endpoint is not needed; 
 \item if no content negotiation during HTTP dereferencing is needed;
\end{itemize}

\paragraph{Pick OpenCyc...}
\begin{itemize}
 \item if especially the representation of classes is important; 
 \item if descriptive URIs are desired; 
 \item if classes should have \texttt{owl:equivalentClass}-relations;  
 \item if a SPARQL endpoint is not needed; 
 \item if no content negotiation during HTTP dereferencing is needed; 
 \item if no HTML representations of KG resoures are needed; 
 \item if many existing instances and classes should have \texttt{owl:sameAs}-links; 
 \item if it is not that important whether linked RDF documents are not accessible any more; 
\end{itemize}

\paragraph{Pick Wikidata...}
\begin{itemize}
 \item if incorrect or missing information should be correctable by the community; 
 \item if the source information per statement is important; 
 \item if it should be possible to model unknown and empty values; 
 \item if the KG should support a ranking of statements; 
 \item if a complete schema (covering all general domains) is important; 
 \item if not only well-known, but also unknown entities should be represented; 
 \item if the KG data should be \textit{continuously} editable and queryable; 
 \item if the period in which statements are valid need to be represented; 
 \item if labels should be available in hundreds of languages; 
 \item if especially non-English labels are needed; 
 \item if n-ary relations are acceptable; 
 \item if external vocabulary should be used to a high degree; 
 \item if \texttt{owl:equivalentClass}-statements to external classes and relations are not that important; 
 \item if instances should be interlinked to DBpedia; 
\end{itemize}

\paragraph{Pick YAGO...}
\begin{itemize}
 \item if syntactic incorrectness in date values due to wildcard usage is acceptable; 
 \item if the source information per statement is important; 
 \item if classes should be linked to WordNet synsets; 
 \item if the period in which statements are valid need to be represented; 
 \item if labels should be available in hundreds of languages; 
 \item if descriptive URIs are desired; 
 \item if instances should be interlinked to DBpedia; 
\end{itemize}

\begin{figure}
    \centering
\begin{mdframed}[roundcorner=10pt]
\textbf{Step 1: Requirement Analysis}
\hrule
\begin{itemize}
    \item Identifying the preselection criteria
    \item Assigning a weight to each data quality criterion
\end{itemize}
\end{mdframed}
\customArrow
\begin{mdframed}[roundcorner=10pt]
\textbf{Step 2: Preselection based on the Preselection Criteria}
\hrule
\begin{itemize}
    \item Manually selecting the KGs that fulfill the preselection criteria
\end{itemize}
\end{mdframed}
\customArrow
\begin{mdframed}[roundcorner=10pt]
\textbf{Step 3: Quantitative Assessment of the KGs}
\hrule
\begin{itemize}
    \item Calculating the data quality metric for each data quality criterion
    \item Calculating the fulfillment degree for each KG, thereby determining the KG with the highest fulfillment degree
\end{itemize}
\end{mdframed}
\customArrow
\begin{mdframed}[roundcorner=10pt]
\textbf{Step 4: Qualitative Assessment of the Result}
\hrule
\begin{itemize}
    \item Assessing the selected KG w.r.t. qualitative aspects
    \item Comparing the selected KG with other preselected KGs
    \end{itemize}
\end{mdframed}
 \caption{Our framework for recommending the most suitable knowledge graph for a given setting.}
  \label{fig:Auswahlverfahren}
\end{figure}

\section{Our Knowledge Graph Recommendation Framework}
\label{sec:framework}

The rule of thumbs presented in the previous section can be considered as simplified heuristics on when to use which KG. In cases in which a more profound investigation concerning the KG choice is indispensable, we can refer to the KG recommendation framework outlined in our survey \cite{FaerberSWJ2016}. The usage of this framework can be summarized as follows:

Given a set of KGs, any person interested in using KGs can use our recommendation framework as shown in Figure~\ref{fig:Auswahlverfahren}. 
In Step~1, the pre-selection criteria regarding KGs and the weights for the single metrics are specified. The pre-selection criteria can be data quality criteria or other criteria and need to be selected based on the use case. The timeliness frequency, i.e., how often the KG is updated, is an example for a data quality criterion. The license under which a KG is provided (e.g., CC0 license) is an example for a general criterion. After weighting the criteria, in Step~2 the KGs which do not fulfill the pre-selection criteria are neglected. 
In Step~3, the fulfillment degrees of the remaining KGs are calculated and the KG with the highest fulfillment degree is selected. Finally, in Step 4 the result can be assessed with regard to qualitative aspects (besides the quantitative assessment performed by means of the data quality metrics) and, if necessary, an alternative KG can be selected for the given scenario.
An example how to use the framework for a concrete use case is given in our article~\cite{FaerberSWJ2016}.

\section{Conclusion}
\label{sec:conclusion}

In this paper, we presented a summary of our work on the data quality of the knowledge graphs DBpedia, Freebase, OpenCyc, Wikidata, and YAGO~\cite{FaerberSWJ2016}. On top of that, we provided simplified rules of thumb on when to use which knowledge graph in a given setting. With these guidelines, you can quickly get a rough idea what the best KGs for your requirements might be.

\bibliographystyle{ACM-Reference-Format}
\bibliography{paper} 


\newcommand{\noop}[1]{}
\begin{thebibliography}{5}


\ifx \showCODEN    \undefined \def \showCODEN     #1{\unskip}     \fi
\ifx \showDOI      \undefined \def \showDOI       #1{#1}\fi
\ifx \showISBNx    \undefined \def \showISBNx     #1{\unskip}     \fi
\ifx \showISBNxiii \undefined \def \showISBNxiii  #1{\unskip}     \fi
\ifx \showISSN     \undefined \def \showISSN      #1{\unskip}     \fi
\ifx \showLCCN     \undefined \def \showLCCN      #1{\unskip}     \fi
\ifx \shownote     \undefined \def \shownote      #1{#1}          \fi
\ifx \showarticletitle \undefined \def \showarticletitle #1{#1}   \fi
\ifx \showURL      \undefined \def \showURL       {\relax}        \fi
\providecommand\bibfield[2]{#2}
\providecommand\bibinfo[2]{#2}
\providecommand\natexlab[1]{#1}
\providecommand\showeprint[2][]{arXiv:#2}

\bibitem[\protect\citeauthoryear{Bizer}{Bizer}{2007}]%
        {Bizer2007}
\bibfield{author}{\bibinfo{person}{Christian Bizer}.}
  \bibinfo{year}{2007}\natexlab{}.
\newblock \bibinfo{booktitle}{{\em {Quality-Driven Information Filtering in the
  Context of Web-Based Information Systems}}}.
\newblock \bibinfo{publisher}{VDM Publishing}.
\newblock


\bibitem[\protect\citeauthoryear{F{\"{a}}rber, Bartscherer, Menne, and
  Rettinger}{F{\"{a}}rber et~al\mbox{.}}{2018}]%
        {FaerberSWJ2016}
\bibfield{author}{\bibinfo{person}{Michael F{\"{a}}rber},
  \bibinfo{person}{Frederic Bartscherer}, \bibinfo{person}{Carsten Menne},
  {and} \bibinfo{person}{Achim Rettinger}.} \bibinfo{year}{2018}\natexlab{}.
\newblock \showarticletitle{{Linked data quality of DBpedia, Freebase, OpenCyc,
  Wikidata, and {YAGO}}}.
\newblock \bibinfo{journal}{{\em Semantic Web\/}} \bibinfo{volume}{9},
  \bibinfo{number}{1} (\bibinfo{year}{2018}), \bibinfo{pages}{77--129}.
\newblock


\bibitem[\protect\citeauthoryear{Markman}{Markman}{2013}]%
        {Markman2013}
\bibfield{author}{\bibinfo{person}{Arthur~B Markman}.}
  \bibinfo{year}{2013}\natexlab{}.
\newblock \bibinfo{booktitle}{{\em {Knowledge representation}}}.
\newblock \bibinfo{publisher}{Psychology Press}.
\newblock


\bibitem[\protect\citeauthoryear{Wang, Reddy, and Kon}{Wang
  et~al\mbox{.}}{1995}]%
        {Wang1995}
\bibfield{author}{\bibinfo{person}{Richard~Y. Wang}, \bibinfo{person}{Martin~P.
  Reddy}, {and} \bibinfo{person}{Henry~B. Kon}.}
  \bibinfo{year}{1995}\natexlab{}.
\newblock \showarticletitle{{Toward quality data: An attribute-based
  approach}}.
\newblock \bibinfo{journal}{{\em Decision Support Systems\/}}
  \bibinfo{volume}{13}, \bibinfo{number}{3} (\bibinfo{year}{1995}),
  \bibinfo{pages}{349--372}.
\newblock


\bibitem[\protect\citeauthoryear{Zaveri, Rula, Maurino, Pietrobon, Lehmann, and
  Auer}{Zaveri et~al\mbox{.}}{2015}]%
        {Zaveri2014}
\bibfield{author}{\bibinfo{person}{Amrapali Zaveri}, \bibinfo{person}{Anisa
  Rula}, \bibinfo{person}{Andrea Maurino}, \bibinfo{person}{Ricardo Pietrobon},
  \bibinfo{person}{Jens Lehmann}, {and} \bibinfo{person}{S{\"o}ren Auer}.}
  \bibinfo{year}{2015}\natexlab{}.
\newblock \showarticletitle{{Quality Assessment for Linked Data: A Survey}}.
\newblock \bibinfo{journal}{{\em Semantic Web\/}} \bibinfo{volume}{7},
  \bibinfo{number}{1} (\bibinfo{year}{2015}), \bibinfo{pages}{63--93}.
\newblock


\end{thebibliography}

\end{document}